\documentclass[conference]{IEEEtran}
\ifCLASSINFOpdf
   \usepackage[pdftex]{graphicx}
\else
   \usepackage[dvips]{graphicx}
\fi

\usepackage{amsmath}
\usepackage{tabularx}
\usepackage{tabulary}
\usepackage{amssymb}
\usepackage{color}
\usepackage{url}
\usepackage{lipsum}
\usepackage{multirow}
\usepackage{algorithm} 
\usepackage{algorithmic}  
\usepackage[algo2e]{algorithm2e}
\newtheorem{definition}{Definition}

\begin{document}
\title{Anomaly Detection in a Digital Video Broadcasting System Using Timed Automata}

\author{\IEEEauthorblockN{Xiaoran Liu\IEEEauthorrefmark{1},
Qin Lin\IEEEauthorrefmark{2}, Sicco Verwer\IEEEauthorrefmark{3} }
\IEEEauthorblockA{Faculty of Electrical Engineering, Mathematics and Computer Science\\
Delft University of Technology\\
Delft, the Netherlands\\
Email: \IEEEauthorrefmark{1}lxran3691@gmail.com,
\IEEEauthorrefmark{2}q.lin@tudelft.nl,
\IEEEauthorrefmark{3}s.e.verwer@tudelft.nl}
\and
\IEEEauthorblockN{Dmitri Jarnikov}
\IEEEauthorblockA{Irdeto B.V., the Netherlands\\
Email: djarnikov@irdeto.com\\
Eindhoven University of Technology\\
Eindhoven, the Netherlands\\}}
\maketitle

\maketitle

\begin{abstract}
This paper focuses on detecting anomalies in a digital video broadcasting (DVB) system from providers' perspective. We learn a probabilistic deterministic real timed automaton profiling benign behavior of encryption control in the DVB control access system. This profile is used as a one-class classifier. Anomalous items in a testing sequence are detected when the sequence is not accepted by the learned model.
\end{abstract}

\IEEEpeerreviewmaketitle

\section{Introduction}
Nowadays proliferation of accessible data stream happens through various networks such as broadcasting, web page, mobile phone, etc. The large volume and continuously real-time updating characteristics of data streams pose challenges for studying and mining. This work focuses on profiling normal behavior from encryption control stream data generated by a digital video broadcasting (DVB) system. From a providers' perspective, such a profile provides insights about the process generating the data and it can be used for anomaly detection.

\subsection{Digital Video Broadcasting system}
The DVB has been adopted in Europe as an open standard for a long time. The DVB standard defines both physical and data link layers for a variety of subsystems, e.g., satellite, cable, terrestrial television, and microwave. Furthermore, the DVB standard formulates how multiple program data are distributed and transported among protocols in format of the MPEG Transport Stream (MPEG-TS). Readers are referred to the official website of the DVB project for more detailed specifications\footnote{\url{https://www.dvb.org/standards}}.

\begin{figure}[h]
\centering
\includegraphics[scale=0.5]{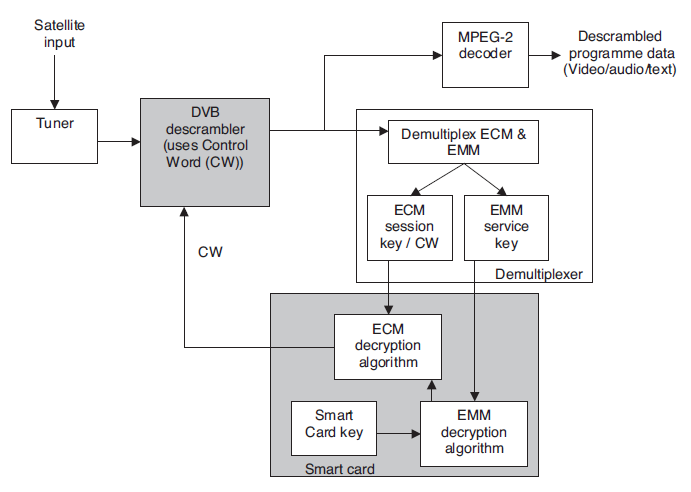}
\caption{Example of the control access system from the consumer's side \protect\cite{PalmieriFioreCastiglioneEtAl2013}.}
\label{fig:cas}
\end{figure}

The objective of our research lies on a core security component of the DVB, i.e., the control access system. Figure \ref{fig:cas} shows how the control access system works from a customer's side.
The TS is received from the satellite through the tuner and arrives at the DVB descrambler. The content of the TS is scrambled with a 48-bit secret key, also known as the control word (CW). The CW is further encrypted into an entitlement control message (ECM). Meanwhile an entitlement management message (EMM) contains the authorization for decrypting the ECM. A demultiplexer extracts the ECM and the EMM from the TS. Valid smart cards are able to decrypt the EMM in order to access the authorization and decrypt the ECM. The decrypted ECM is used to descramble the TS into MPEG-2. The MPEG-2 contains video/audio/context, which is transported to users.

\subsection{Related Work}
Anomalies are usually considered as outliers, surprises, exceptions, noises, and novelties \cite{chandola2009anomaly}. Anomaly detection can be achieved by many pattern recognition techniques. However, in many domains, e.g., medical and financial area, frauds or anomalies detection should be understandable and interpretable. A key reason being that the learned knowledge should be validated by humans before being used. We provide a brief literature review of anomaly detection using language models since they are one of the most understandable models. Lin and et al. represent time series using the Symbolic Aggregate approXimation (SAX) \cite{lin2007experiencing}, and then use a context-free grammar for rules' generation \cite{senin2015time}. Intuitively, anomalies tend to have low grammar density (novelty patterns are less compressed). Sekar and et al. learn program behaviors from system call sequences using a finite-state automaton for intrusion detection \cite{sekar2001fast}. Timo and et al. propose a tailored behavior model (a probabilistic deterministic timed-transition automaton) to identify anomalies \cite{klerx2014model}. They first augment the learned automaton with timing information, after which traces of ATM observations are traversed by the model. Aggregated transition probabilities are then compared with a predetermined threshold for detecting anomalies.

In this paper, we study event sequences from a DVB system. These sequences contain the encryption information needed for viewing the stream content. We model these sequences using PDRTAs (Probabilistic Deterministic Real Timed Automata), which we learn unsupervised from a set of input sequences using the RTI+ algorithm (Real-Time Identification from Positive Data)~\cite{VerwerWeerdtWitteveen2010}. first, we compute the time difference between two consecutive and distinct events to obtain timed strings.
We then segment the time strings
into frames. The sequences are fed into the RTI+ algorithm to learn a timed automaton. In the testing phrase, an anomaly of a sequence is identified if the sequence is not accepted by the learned model. This paper makes the following contributions:
\begin{itemize}
\item To the best of our knowledge, this paper is the first one using a timed automaton to detect anomalies in the DVB system. Two types of anomalies, i.e., \emph{data lost} and \emph{timing error}, are identified with low false positive rate.
\item The model provides highly interpretable insights for understanding the underlying process generating the data. Experts from the DVB area can easily monitor and validate the system under operation using such a model.
\end{itemize}
This paper is organized as follows. Section \ref{sec:des} introduces the  data preprocessing. Section \ref{sec:method} discusses the learning algorithm and the experimental results. We make concluding remarks in Section \ref{sec:conclusion}. 

\section{Data Description}
\label{sec:des}
An encryption scheme sequence (ESS) is abstracted from the monitoring records of the MPEG-TS. It is essentially a discrete event sequence of ECM streaming in the control access system.

The alphabet of the ESS is $\{A, B, 0, 1\}$. The symbols $A$ and $B$ abstractly stand for two ECMs containing an even and an odd key, respectively. The symbols $0$ and $1$ are even and odd encryption modes. According to expert knowledge,
some known rules are the following:
\begin{enumerate}
\item An ESS starts from $A$s, followed by $1$s, $B$s, then ends with $0$s. A complete encryption scheme consists of two aforementioned sequences.
\item The number of symbols are not fixed in an ESS. However, the amount of $A$s is always equal to that of $B$s. So are the paired events of 1s and 0s.
\end{enumerate}
We can use a simple regular expression to generate the aforementioned rules: $(A\{m\}1\{n\}B\{m\}0\{n\})\{2\}$, where $m$ and $n$ are the legitimate numbers for an ECM pair (A/B) and an encryption mode pair (1/0). Basically there are two kinds of anomalies in practice of the DVB. The first one is \emph{data lost} and the other one is \emph{timing error}, e.g., too large or too small time delay.

\section{Methods and Results}
In this section, we will deploy a language model, i.e., a timed automaton for anomaly detection. Timed automata \emph{explicitly} model the underlying varying-duration behavior of ESS streaming.
\label{sec:method}
\subsection{RTI+}
Time constrains are $implicit$ in conventional discrete event systems, e.g., $n$-grams or hidden Markov models. However, time information is often important for modeling the behavior of such systems. An algorithm for efficient learning of timed automata algorithm named RTI+ (Real-Time Identification from Positive Data) was proposed by~\cite{VerwerWeerdtWitteveen2010}. Discrete events are represented by timed strings $(a_1,t_1)\allowbreak(a_2,t_2)\allowbreak\cdots\allowbreak(a_n,t_n)$, where $a_i$ is a discrete event occurring with $t_i$ time delay since the $i-1$th event. A PDRTA (probabilistic deterministic real timed automaton) model defines a probability distribution over such timed strings, having a Markov property in the distribution over events, and a semi-Markov property in the time guards. 
\begin{definition}
A probabilistic DRTA (PDRTA) $A$ is a quadruple $\langle A',H,S,T\rangle$, where $A'=\langle Q,\Sigma,\Delta,q_0\rangle$ is a DRTA without final states, $H$ is a finite set of bins (time intervals) $[v,v'], v, v'\in\mathbb{N}$, known as the histogram, $S$ is a finite set of symbol probability distributions $S_q={Pr(S=a\mid q)\mid a\in \Sigma, q\in Q}$, and $T$ is a finite set of time-bin probability distributions $T_q ={Pr(T\in h\mid q)\mid h\in H,q\in Q}$.
\end{definition}
In a PDRTA, the state transition is triggered when both the event and the time guard are satisfied.

Table \ref{tab:data} is an example of how to obtain a timed string. Raw data is formatted in tuples of a symbol and a timestamp. In timed strings, time delay between events represents event transition intervals. Note that since only integer format time is readable for RTI+, the original float time with precision microseconds need to be rounded. An interesting problem what precision (different level of magnification) is ``optimal'' when learning the time guards. A large magnification intuitively makes time guards more sensitive. We will conduct several comparative trials to find a good precision in the experimental part.
\begin{table*}[ht]
\centering
\caption{Data Preprocessing}
\begin{tabular}{c|cccccccccc}
\hline
Original Data&$A, 0$& $A, 1$& $A, 3$ & $1, 13$ & $1, 13$ &$B, 15$& $B, 15$ & $B, 16$ & $0, 16$ & $0, 17$\\
\hline
Time String&$(A,0)$&$(A,1)$&$(A, 2)$&$(1,10)$&$(1,0)$&$(B,2)$&$(B,0)$&$(B,1)$ & $(0, 0)$ & $(0, 1)$\\
\hline
\end{tabular}
\label{tab:data}
\end{table*}
\begin{algorithm}[hbt]
\caption{Data identification with RTI+:\label{algo:RTI+}}
\KwIn{A (multi-)set of timed strings $S_{+}$}
\KwOut{A small PDRTA $\mathcal{A}$ for $S_{+}$}
Construct a timed prefix $\mathcal{A}$ tree from $S_{+}$, let $Q' = \emptyset$;\\
\For{ all transitions $\delta=\langle{q,q',a,[m,m^{\prime}]}\rangle$ from $\mathcal{A}$,}
{
Evaluate all possible merges of $q'$ with states from $Q'$;

Evaluate all possible splits of $\delta$;

\If{the lowest split $ ${p-value}$ < 0.05$}
{
perform this split;
}
\ElseIf{the highest merge $ ${p-value}$ > 0.05$}
{
perform this merge;
}
\Else
{
add $q$ to $Q'$;
}
}
\end{algorithm}
\subsection{model interpretation}
\begin{figure*}[htbp]
\centering
\includegraphics[width=180mm]{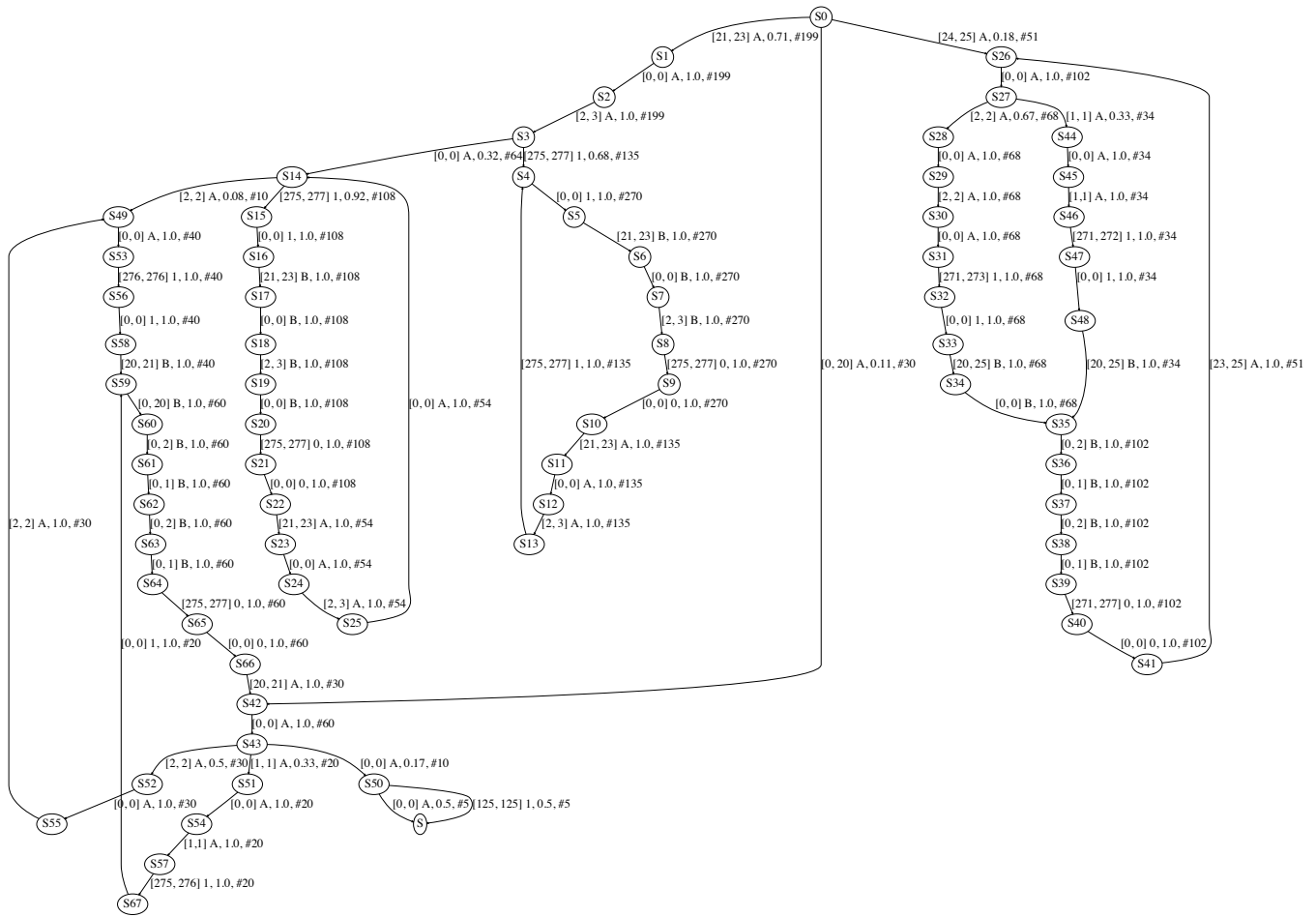}
\caption{Experimental process based on RTI+}
\label{fig:10}
\end{figure*}

We print out the automaton learned using RTI+ from the DVB data in Figure \ref{fig:10}. All the states are depicted using circles. The arcs represent transitions between the states. A transition is triggered when both an event and its timing are valid (inside a time guard). The event and the timing, along with their probability and occurrence, are printed next to the arc. The sink state named $S$ is generated due to some strings with very low occurrence (less then 5). Some interesting knowledge is discovered in such a insightful and structural model.
\begin{enumerate}
\item {Loops indicate the cycling behavior of sending ESSs. For example the path S0-S1-S2-S3-S4-S5-...S4-S5 can be expressed as: $(A\{3\}1\{2\}B\{3\}0\{2\})+$.}
\item{Without considering the time information, the legitimate rule can be drawn as $(A\{3\textnormal{-}6\}1\{2\}B\{3\textnormal{-}6\}0\{2\})+$.}
\item{RTI+ splits at some states, e.g., S0 for distinct time guards: $[0, 20]$, $[21, 23]$, and $[24, 25]$.} RTI+ splits the time in a transition to pull apart distinguished tails. Take the two sequences sharing same symbolic rules $(A\{5\}1\{2\}B\{5\}0\{2\})+$: S0-S26-S27-S44-S45-S46-...S41-S26 and S0-S42-S43-S51-...S59-S60 for instance, they are distinguished as \emph{low} and \emph{high} delay behaviors (loops) of sending the first ECM key: $A$. This type of information indicates the possible varying-duration properties of the system, which deserves a further verification by the DVB experts.
\end{enumerate}

\subsection{result}
The training set contains 280 sequences in total. In the testing phrase, we collect two data sets having a data lost and a timing error, respectively. The number of sequence in each of them is 29. Anomalous items in a sequence will be reported if such a sequence is not accepted by this model. The Table \ref{tab:misstest} and \ref{tab:timetest} show the results of data lost and time delay testing data set. True positive (TP), false negative (FN), false positive (FP), and true negative (TN) are listed in the tables. Their corresponding rates i.e., TPR, FNR, FPR, TNR, and accuracy (ACC) are computed as well.

We compare three levels of magnification $10$, $10^3$, and $10^6$ for determining good time precision. The results show that the precision with $10^6$ times scaling performs best. We suggest to well address this problem by comparative trials in practice. The false positive is caused by the sink state in our model, i.e., because RTI+ does not handle transitions that occur very infrequently. The sequences firing these transitions provide too little information for RTI+ to obtain reliable statistical tests for computing merge and split scores. To avoid this problem, we suggest to check strings in the training data as well as the model. If it does exist in the training data, even though it traverses to a sink state, we can label it as normal behavior to reduce the false positives.

Two anomalous sequences from the DVB system are listed as follows.
\begin{itemize}
\item{Data lost:}(1, 300), (1, 0), (B, 22), (B, 0), (B, 2), (B, 0), (0, 276), (0, 0), (A, 22), (A, 0), (A, 2), (A, 0), (1, 276), (1, 0), (B, 22), (B, 0), (B, 2), (B, 0), (0, 276), (0, 0) 
\item{Time delay:} {(A, 22), (A, 0), (A, 2), (A, 275), (1, 0), (1, 0), (B, 22), (B, 0), (B, 2), (B, 0), (0, 275), (0, 0), (A, 22), (A, 0), (A, 2), (A, 0), (1, 275), (1, 0), (B, 22), (B, 0), (B, 3), (B, 0), (0, 275), (0, 0)}
\end{itemize}
The first sequence misses $A$s in the beginning, which leads to a failure triggering the following transitions at the start state. TIn the second sequence, the 4th $A$ is outside the valid time guard $[0, 0]$.
\begin{table}[h]
\centering
\begin{tabular}{c c c c c c c c c c}
\hline
&TP&FN&FP&TN&TPR&FNR&FPR&TNR&ACC\\
\hline
$10$ &1&0&1&27&1.00&0.00&0.04&0.96&0.97 \\
$10^3$ &1&0&6&22&1.00&0.00&0.21&0.79&0.79 \\
$10^6$ &1&0&0&28&1.00&0.00&0.00&1.00&1.00 \\
\hline
\end{tabular}
\caption{the result of three automata when testing data lost anomaly}
\label{tab:misstest}
\end{table}
\begin{table}[htbp!]
\centering
\begin{tabular}{c c c c c c c c c c}
\hline
&TP&FN&FP&TN&TPR&FNR&FPR&TNR&ACC\\
\hline
$10$&1&0&1&27&1.00&0.00&0.04&0.96&0.96\\
$10^3$&1&0&5&23&1.00&0.00&0.18&0.82&0.83\\
$10^6$&1&0&1&27&1.00&0.00&0.04&0.96&0.96\\
\hline
\end{tabular}
\caption{the result of three automata when testing time delay anomaly}
\label{tab:timetest}
\end{table}

\section{Conclusion}
\label{sec:conclusion}
We learn a benign behavioral profile using RTI+. Such a generative model provides insights for the ESSs screaming process. The knowledge we discover is consistent with that of the experts from the DVB system area. In addition, our model provides the valid timing in each transition for behavior duration verification. The experiments demonstrate that our model has high accuracy in anomaly detection. Another advantage of our model is that it is efficient for real-time application because the verification is just firing transitions in our model, which is polynomial in time.
\section*{Acknowledgment}
This work is partially supported by Technologiestichting STW VENI project 13136 (MANTA) and NWO project 62001628 (LEMMA).
\bibliographystyle{./IEEEtran}
\bibliography{./IEEEabrv,./main}

\begin{thebibliography}{1}
\providecommand{\url}[1]{#1}
\csname url@samestyle\endcsname
\providecommand{\newblock}{\relax}
\providecommand{\bibinfo}[2]{#2}
\providecommand{\BIBentrySTDinterwordspacing}{\spaceskip=0pt\relax}
\providecommand{\BIBentryALTinterwordstretchfactor}{4}
\providecommand{\BIBentryALTinterwordspacing}{\spaceskip=\fontdimen2\font plus
\BIBentryALTinterwordstretchfactor\fontdimen3\font minus
  \fontdimen4\font\relax}
\providecommand{\BIBforeignlanguage}[2]{{%
\expandafter\ifx\csname l@#1\endcsname\relax
\typeout{** WARNING: IEEEtran.bst: No hyphenation pattern has been}%
\typeout{** loaded for the language `#1'. Using the pattern for}%
\typeout{** the default language instead.}%
\else
\language=\csname l@#1\endcsname
\fi
#2}}
\providecommand{\BIBdecl}{\relax}
\BIBdecl

\bibitem{PalmieriFioreCastiglioneEtAl2013}
F.~Palmieri, U.~Fiore, A.~Castiglione, and A.~De~Santis, ``On the detection of
  card-sharing traffic through wavelet analysis and support vector machines,''
  \emph{Applied Soft Computing}, vol.~13, no.~1, pp. 615--627, 2013.

\bibitem{chandola2009anomaly}
V.~Chandola, A.~Banerjee, and V.~Kumar, ``Anomaly detection: A survey,''
  \emph{ACM computing surveys (CSUR)}, vol.~41, no.~3, p.~15, 2009.

\bibitem{lin2007experiencing}
J.~Lin, E.~Keogh, L.~Wei, and S.~Lonardi, ``Experiencing sax: a novel symbolic
  representation of time series,'' \emph{Data Mining and Knowledge Discovery},
  vol.~15, no.~2, pp. 107--144, 2007.

\bibitem{senin2015time}
P.~Senin, J.~Lin, X.~Wang, T.~Oates, S.~Gandhi, A.~P. Boedihardjo, C.~Chen, and
  S.~Frankenstein, ``Time series anomaly discovery with grammar-based
  compression.'' in \emph{EDBT}, 2015, pp. 481--492.

\bibitem{sekar2001fast}
R.~Sekar, M.~Bendre, D.~Dhurjati, and P.~Bollineni, ``A fast automaton-based
  method for detecting anomalous program behaviors,'' in \emph{Security and
  Privacy, 2001. S\&P 2001. Proceedings. 2001 IEEE Symposium on}.\hskip 1em
  plus 0.5em minus 0.4em\relax IEEE, 2001, pp. 144--155.

\bibitem{klerx2014model}
T.~Klerx, M.~Anderka, H.~K. B{\"u}ning, and S.~Priesterjahn, ``Model-based
  anomaly detection for discrete event systems,'' in \emph{Tools with
  Artificial Intelligence (ICTAI), 2014 IEEE 26th International Conference
  on}.\hskip 1em plus 0.5em minus 0.4em\relax IEEE, 2014, pp. 665--672.

\bibitem{VerwerWeerdtWitteveen2010}
S.~Verwer, M.~de~Weerdt, and C.~Witteveen, ``A likelihood-ratio test for
  identifying probabilistic deterministic real-time automata from positive
  data,'' in \emph{International Colloquium on Grammatical Inference}.\hskip
  1em plus 0.5em minus 0.4em\relax Springer, 2010, pp. 203--216.

\end{thebibliography}
\end{document}